\documentclass[11pt,a4paper]{article}
\usepackage[hyperref]{emnlp2018}
\usepackage{times}
\usepackage{latexsym}

\usepackage{url}

\usepackage{amsfonts}
\usepackage{amsmath}

\DeclareMathOperator*{\argmax}{argmax}

\usepackage{tikz}
\usetikzlibrary{arrows,shapes,trees}
\usepackage{amsmath}
\usepackage{amssymb}
\usepackage{enumitem}

\aclfinalcopy 

\title{Neural Latent Extractive Document Summarization}

\author{Xingxing Zhang$^\dag$, Mirella Lapata$^\ddag$, Furu Wei$^\dag$ \and Ming Zhou$^\dag$ \\
	$^\dag$Microsoft Research Asia, Beijing, China \\
	$^\ddag$Institute for Language, Cognition and Computation, \\ School of Informatics, University of Edinburgh, UK \\ 
	{\tt \{xizhang,fuwei,mingzhou\}@microsoft.com,mlap@inf.ed.ac.uk} }

\date{}

\begin{document}
\maketitle
\begin{abstract}
  Extractive summarization models require sentence-level labels, which
  are usually created heuristically (e.g., with rule-based methods)
  given that most summarization datasets only have document-summary
  pairs. Since these labels might be suboptimal, we propose a latent
  variable extractive model where sentences are viewed as latent
  variables and sentences with activated variables are used to infer
  gold summaries. During training the loss comes \emph{directly} from
  gold summaries. Experiments on the CNN/Dailymail dataset show that
  our model improves over a strong extractive baseline trained on
  heuristically approximated labels and also performs competitively to
  several recent models.
\end{abstract}

\section{Introduction}

Document summarization aims to automatically rewrite a document into a
shorter version while retaining its most important content.  Of the
many summarization paradigms that have been identified over the years
(see \citealt{Mani:01} and \citealt{Nenkova:McKeown:2011} for
comprehensive overviews), two have consistently attracted attention:
\emph{extractive} approaches generate summaries by {copying} parts of
the source document (usually whole sentences), while
\emph{abstractive} methods may generate new words or phrases which are not
in the document.

A great deal of previous work has focused on extractive summarization
which is usually modeled as a sentence ranking or binary
classification problem (i.e.,~sentences which are top ranked or
predicted as {\tt True} are selected as summaries). Early attempts
mostly leverage human-engineered features
\cite{filatova:2004:acl:workshop} coupled with binary classifiers
\cite{kupiec:1995:sigir}, hidden Markov models
\cite{conroy:2001:sigir}, graph based methods
\cite{mihalcea:2005:acl}, and integer linear programming
\cite{woodsend:2010:acl}.

The successful application of neural network models to a variety of
NLP tasks and the availability of large scale summarization datasets
\cite{hermann:2015:nips,nallapati:2016:arxiv} has provided strong
impetus to develop data-driven approaches which take advantage of
continuous-space representations.  \newcite{cheng:2016:acl} propose a
hierarchical long short-term memory network (LSTM;
\citealt{hochreiter:1997:nc}) to learn context dependent sentence
representations for a document and then use yet another LSTM decoder
to predict a binary label for each
sentence. \newcite{nallapati:2017:aaai} adopt a similar approach, they
differ in their neural architecture for sentence encoding and the
features used during label prediction, while \citet{Narayan:ea:2018}
equip the same architecture with a training algorithm based on
reinforcement learning.  Abstractive models
\cite{nallapati:2016:arxiv,see:2017:acl,paulus:2017:arxiv} are based
on sequence-to-sequence learning
\cite{sutskever:2014:nips,bahdanau:2015:iclr}, however, most of them
underperform or are on par with the baseline of simply selecting the
leading sentences in the document as summaries (but see
\citealt{paulus:2017:arxiv} and \citealt{N18-1150} for exceptions).



Although seemingly more successful than their abstractive
counterparts, extractive models require sentence-level labels, which
are not included in most summarization datasets (only document and
gold summary pairs are available). Sentence labels are usually
obtained by rule-based methods \cite{cheng:2016:acl} or by maximizing
the \textsc{Rouge} score \cite{lin:2004:acl:w} between a subset of
sentences and the human written summaries
\cite{nallapati:2017:aaai}. These methods do not fully exploit the
human summaries, they only create {\tt True}/{\tt False} labels which
might be suboptimal.  In this paper we propose a latent variable
extractive model and view labels of sentences in a document as binary
latent variables (i.e., zeros and ones). Instead of maximizing the
likelihood of ``gold'' standard labels, the latent model directly
maximizes the likelihood of human summaries given selected sentences.
Experiments on the CNN/Dailymail dataset \cite{hermann:2015:nips} show
that our latent extractive model improves upon a strong extractive
baseline trained on rule-based labels and also performs competitively
to several recent models.

%
%

%

\section{Model}
We first introduce the neural extractive summarization model upon
which our latent model is based on. We then describe a sentence
compression model which is used in our latent model and finally move
on to present the latent model itself.

\begin{figure}[t]
	
	\centering
	\resizebox{0.48\textwidth}{!}{%
		
		\centering
		\begin{tikzpicture}[scale=.75,->,>=stealth',thick,main node/.style={rectangle,rounded corners=3pt,fill=blue!10,draw,font=\sffamily\Large\bfseries,inner sep=0,minimum size=2.5mm,minimum width=3mm,minimum height=1cm,path picture={
				\draw[fill=blue!50!black] (0, -0.25) circle (0.8mm);
				\draw[fill=blue!50!black] (0, 0.25) circle (0.8mm);
			}
		},
		word node/.style={rectangle,rounded corners=3pt,fill=red!30!white,draw,font=\sffamily\bfseries,inner sep=3pt,minimum size=2.5mm}
		]
		
		\node[align=center] (xh_0) at (-4.7, 0) {\bf Document \\ \bf Encoder};
		\node[main node] (xh_1) at (-2, 0) {};
		\node[main node] (xh_2) at (0, 0) {};
		\node[main node] (xh_3) at (2, 0) {};
		\node[main node] (xh_4) at (4, 0) {};
		\node (xh_5) at (5.5, 0) {};
		
		\draw[line width=2pt,blue] (xh_0) -- (xh_1);
		\draw[line width=2pt,blue] (xh_5) -- (xh_4);
		
		\path (xh_1.east) -- (xh_1.north east) coordinate[pos=0.5] (xh_1_0);
		\path (xh_2.west) -- (xh_2.north west) coordinate[pos=0.5] (xh_2_0);
		\draw[line width=2pt,blue] (xh_1_0) -- (xh_2_0);
		
		\path (xh_1.east) -- (xh_1.south east) coordinate[pos=0.5] (xh_1_1);
		\path (xh_2.west) -- (xh_2.south west) coordinate[pos=0.5] (xh_2_1);
		\draw[line width=2pt,blue] (xh_2_1) -- (xh_1_1);
		
		\path (xh_2.east) -- (xh_2.north east) coordinate[pos=0.5] (xh_2_0);
		\path (xh_3.west) -- (xh_3.north west) coordinate[pos=0.5] (xh_3_0);
		\draw[line width=2pt,blue] (xh_2_0) -- (xh_3_0);
		
		\path (xh_2.east) -- (xh_2.south east) coordinate[pos=0.5] (xh_2_1);
		\path (xh_3.west) -- (xh_3.south west) coordinate[pos=0.5] (xh_3_1);
		\draw[line width=2pt,blue] (xh_3_1) -- (xh_2_1);
		
		\path (xh_3.east) -- (xh_3.north east) coordinate[pos=0.5] (xh_3_0);
		\path (xh_4.west) -- (xh_4.north west) coordinate[pos=0.5] (xh_4_0);
		\draw[line width=2pt,blue] (xh_3_0) -- (xh_4_0);
		
		\path (xh_3.east) -- (xh_3.south east) coordinate[pos=0.5] (xh_3_1);
		\path (xh_4.west) -- (xh_4.south west) coordinate[pos=0.5] (xh_4_1);
		\draw[line width=2pt,blue] (xh_4_1) -- (xh_3_1);

		\def\sentheight{-2}
		\node (sent_1) at (-2, \sentheight) {\Large $\text{sent}_1$};
		\node (sent_2) at (0, \sentheight) {\Large $\text{sent}_2$};
		\node (sent_3) at (2, \sentheight) {\Large $\text{sent}_3$};
		\node (sent_4) at (4, \sentheight) {\Large $\text{sent}_4$};

		\draw[line width=2pt,brown] (sent_1) -- (xh_1);
		\draw[line width=2pt,brown] (sent_2) -- (xh_2);
		\draw[line width=2pt,brown] (sent_3) -- (xh_3);
		\draw[line width=2pt,brown] (sent_4) -- (xh_4);

		\def\decheight{2}
		\node[align=center] (yh_0) at (-4.7, \decheight) {\bf Document \\ \bf Decoder};
		\node[main node] (yh_1) at (-2, \decheight) {};
		\node[main node] (yh_2) at (0, \decheight) {};
		\node[main node] (yh_3) at (2, \decheight) {};
		\node[main node] (yh_4) at (4, \decheight) {};
		
		\draw[line width=1.5pt,red] (xh_1) -- (yh_1);
		\draw[line width=1.5pt,red] (xh_2) -- (yh_2);
		\draw[line width=1.5pt,red] (xh_3) -- (yh_3);
		\draw[line width=1.5pt,red] (xh_4) -- (yh_4);
		
		\draw[line width=2pt,black] (yh_1) -- (yh_2);
		\draw[line width=2pt,black] (yh_2) -- (yh_3);
		\draw[line width=2pt,black] (yh_3) -- (yh_4);
		\draw[line width=2pt,black] (yh_0) -- (yh_1);
		
		\def\yheight{4}
		\node[align=center] (y0) at (-4.7, \yheight) {Latent \\ Variables};
		\node[circle,fill=red!60!white] (y1) at (-2, \yheight) {1};
		\node[circle,fill=yellow!60!white] (y2) at (0, \yheight) {0};
		\node[circle,fill=red!60!white] (y3) at (2, \yheight) {1};
		\node[circle,fill=yellow!60!white] (y4) at (4, \yheight) {0};
		
		\draw[line width=2pt,black] (yh_1) -- (y1);
		\draw[line width=2pt,brown] (y1) -- (yh_2);
		\draw[line width=2pt,black] (yh_2) -- (y2);
		\draw[line width=2pt,brown] (y2) -- (yh_3);
		\draw[line width=2pt,black] (yh_3) -- (y3);
		\draw[line width=2pt,brown] (y3) -- (yh_4);
		\draw[line width=2pt,black] (yh_4) -- (y4);
		
		\def\yheight{5.8}
		\node[rectangle,rounded corners=3pt,fill=blue!25!white] (sum1) at (-0.5, \yheight) {$\text{sum\_sent}_1$};
		\node[rectangle,rounded corners=3pt,fill=blue!25!white] (sum2) at (2.6, \yheight) {$\text{sum\_sent}_2$};
		
		\draw[line width=1.5pt,red] (y1) -- (sum1);
		\draw[line width=1.5pt,blue!50!black] (y3) -- (sum1);
		
		\draw[line width=1.5pt,blue!50!black] (y1) -- (sum2);
		\draw[line width=1.5pt,red] (y3) -- (sum2);

		\def\wordheight{-4}
		
		\node[align=center] (wh_m1) at (-4.7, \wordheight) {\bf Sentence \\ \bf Encoder};
		\node[align=center] (wh_0) at (-1.5, \wordheight) {};
		\node[main node] (wh_1) at (0, \wordheight) {};
		\node[main node] (wh_2) at (2, \wordheight) {};
		\node[main node] (wh_3) at (4, \wordheight) {};
		\node (wh_5) at (5.5, \wordheight) {};
		
		\def\wordheight{-5.5}
		\node (w_0) at (-2, \wordheight) {\Large $\text{sent}_3=$};
		\node (w_1) at (0, \wordheight) {\Large $w_1$};
		\node (w_2) at (2, \wordheight) {\Large $w_2$};
		\node (w_3) at (4, \wordheight) {\Large $w_3$};
		
		\path (wh_1.east) -- (wh_1.north east) coordinate[pos=0.5] (wh_1_0);
		\path (wh_2.west) -- (wh_2.north west) coordinate[pos=0.5] (wh_2_0);
		\draw[line width=2pt,blue] (wh_1_0) -- (wh_2_0);
		
		\path (wh_1.east) -- (wh_1.south east) coordinate[pos=0.5] (wh_1_1);
		\path (wh_2.west) -- (wh_2.south west) coordinate[pos=0.5] (wh_2_1);
		\draw[line width=2pt,blue] (wh_2_1) -- (wh_1_1);
		
		\path (wh_2.east) -- (wh_2.north east) coordinate[pos=0.5] (wh_2_0);
		\path (wh_3.west) -- (wh_3.north west) coordinate[pos=0.5] (wh_3_0);
		\draw[line width=2pt,blue] (wh_2_0) -- (wh_3_0);
		
		\path (wh_2.east) -- (wh_2.south east) coordinate[pos=0.5] (wh_2_1);
		\path (wh_3.west) -- (wh_3.south west) coordinate[pos=0.5] (wh_3_1);
		\draw[line width=2pt,blue] (wh_3_1) -- (wh_2_1);

		\draw[line width=2pt,blue] (wh_0) -- (wh_1);
		\draw[line width=2pt,blue] (wh_5) -- (wh_3);
		
		\draw[line width=1pt,brown] (w_1) -- (wh_1);
		\draw[line width=1pt,brown] (w_2) -- (wh_2);
		\draw[line width=1pt,brown] (w_3) -- (wh_3);

		\node [trapezium, trapezium angle=30, minimum width=23mm, draw, thick,
		inner xsep=3pt, inner ysep=3pt,fill=purple!20] at (2,-2.8) {MeanPool};

		\end{tikzpicture}
		
	}%
	\vspace*{-.3cm}	
	\caption{Latent variable extractive summarization
          model. $\text{sent}_i$ is a sentence in a document and
          $\text{sum\_sent}_i$ is a sentence in a gold summary of the
          document.}
	\label{fig:model-overview}
\end{figure}
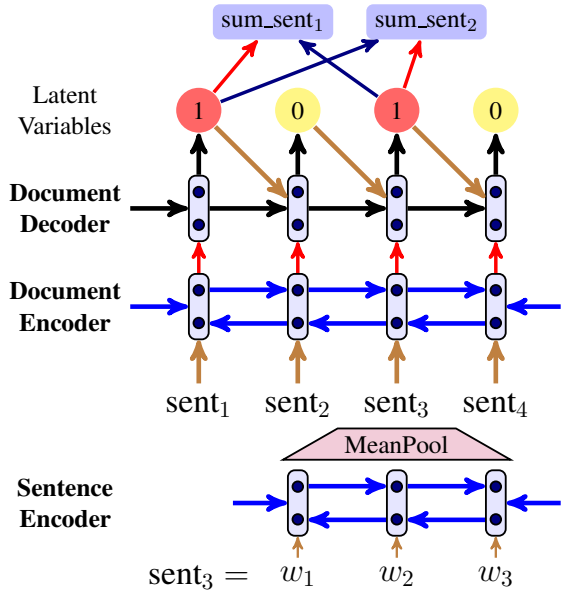

\subsection{Neural Extractive Summarization}
\label{sec:extract_sum}
In extractive summarization, a subset of sentences in a document is
selected as its summary. We model this problem as an instance of
sequence labeling. Specifically, a document is viewed as a sequence of
sentences and the model is expected to predict a {\tt True} or {\tt
  False} label for each sentence, where {\tt True} indicates that the
sentence should be included in the summary. It is assumed that during
training sentences and their labels in each document are given
(methods for obtaining these labels are discussed in
Section~\ref{sec:exp}).

As shown in the lower part of Figure~\ref{fig:model-overview}, our
extractive model has three parts: a \emph{sentence encoder} to convert
each sentence into a vector, a \emph{document encoder} to learn
sentence representations given surrounding sentences as context, and a
\emph{document decoder} to predict sentence labels based on
representations learned by the {document encoder}. Let
$\mathcal{D}=(S_1, S_2, \dots, S_{|\mathcal{D}|})$ denote a document
and $S_i=(w_1^i, w_2^i, \dots, w_{|S_i|}^i)$ a sentence in
$\mathcal{D}$ (where $w_j^i$ is a word in $S_i$). Let $Y=(y_1, \dots,
y_{|\mathcal{D}|})$ denote sentence labels. The {sentence encoder}
first transforms $S_i$ into a list of hidden states $(\mathbf{h}_1^i,
\mathbf{h}_2^i, \dots, \mathbf{h}_{|S_i|}^i)$ using a Bidirectional
Long Short-Term Memory Network (Bi-LSTM;
\citealt{hochreiter:1997:nc,schuster:1997:tsp}). Then, the {sentence
  encoder} yields~$\mathbf{v}_i$, the representation of $S_i$, by
averaging these hidden states (also see
Figure~\ref{fig:model-overview}):
\begin{equation}
\mathbf{v}_i = \frac{1}{|S_i|} \sum_j \mathbf{h}_j^i
\end{equation}

In analogy to the {sentence encoder}, the {document encoder} is
another Bi-LSTM but applies on the sentence level. After running the
Bi-LSTM on a sequence of sentence representations $(\mathbf{v}_1,
\mathbf{v}_2, \dots, \mathbf{v}_{|\mathcal{D}|})$, we obtain 
context dependent sentence representations $(\mathbf{h}^{E}_1,
\mathbf{h}^E_2, \dots, \mathbf{h}^E_{|\mathcal{D}|})$.

The document decoder is also an LSTM which predicts sentence
labels. At each time step, it takes the context dependent sentence
representation of~$S_i$ produced by the document encoder as
well as the prediction in the previous time step:
\begin{equation}
\mathbf{h}^D_i = \text{LSTM}(\mathbf{h}^D_{i-1}, \begin{bmatrix}
\mathbf{W}_{e} \,\, e(y_{i-1}) \\
\mathbf{h}^E_i
\end{bmatrix})
\end{equation}
where $\mathbf{W}_{e} \in \mathbb{R}^{d \times 2}$ is the label
embedding matrix ($d$ is the hidden dimension for the document decoder
LSTM) and $y_{i-1}$ is the prediction at time step~$i-1$; the
predicted label distribution for $y_i$ is:
\begin{equation}
\label{eq:lbl_dist}
p(y_i|y_{1:i-1}, \mathbf{h}^D_{i-1}) = \text{softmax}(\mathbf{W}_o \, \mathbf{h}^D_i)
\end{equation}
where $\mathbf{W}_{o} \in \mathbb{R}^{2 \times d}$.

The model described above is usually trained by minimizing the
negative log-likelihood of sentence labels in training documents; it
is almost identical to \newcite{cheng:2016:acl} except that we use a
word-level long short-term memory network coupled with mean pooling to
learn sentence representations, while they use convolutional neural
network coupled with max pooling \cite{kim:2016:aaai}.

\subsection{Sentence Compression}
\label{sec:compress}

We train a sentence compression model to map a sentence selected by
the extractive model to a sentence in the summary. The model can be
used to evaluate the quality of a selected sentence with respect to
the summary (i.e.,~the degree to which it is similar) or rewrite an
extracted sentence according to the style of the summary. 

For our compression model we adopt a standard attention-based
sequence-to-sequence architecture
\cite{bahdanau:2015:iclr,rush:2015:emnlp}. The training set for this
model is generated from the same summarization dataset used to train
the exractive model.  Let $\mathcal{D}=(S_1, S_2, \dots,
S_{|\mathcal{D}|})$ denote a document and $\mathcal{H}=(H_1, H_2,
\dots, H_{|\mathcal{H}|})$ its summary. We view each sentence $H_i$ in
the summary as a target sentence and assume that its corresponding
source is a sentence in $\mathcal{D}$ most similar to it. We measure
the similarity between source sentences and candidate targets using
\textsc{Rouge}, i.e.,~$S_j=\argmax_{S_j} \text{ROUGE}(S_j, H_i)$ and
$\langle S_j, H_i \rangle$ is a training instance for the compression
model. The probability of a sentence $\hat{H_i}$ being the compression
of $\hat{S_j}$ (i.e.,~$p_{s2s}(\hat{H_i} | \hat{S_j})$) can be
estimated with a trained compression model.

\subsection{Latent Extractive Summarization} 
\label{sec:latent}
Training the extractive model described in
Section~\ref{sec:extract_sum} requires sentence-level labels which are
obtained heuristically \cite{cheng:2016:acl,nallapati:2017:aaai}.  Our
latent variable model views sentences in a document as binary
variables (i.e.,~zeros and ones) and uses sentences with activated
latent variables (i.e., ones) to infer gold summaries. The latent
variables are predicted with an extractive model and the loss during
training comes from gold summaries \emph{directly}.


Let $\mathcal{D}=(S_1, S_2, \dots, S_{|\mathcal{D}|})$ denote a
document and $\mathcal{H}=(H_1, H_2, \dots, H_{|\mathcal{H}|})$ its
human summary ($H_k$ is a sentence in $\mathcal{H}$). We assume that
there is a latent variable $z_i \in \{0, 1\}$ for each sentence $S_i$
indicating whether $S_i$ should be selected, and~$z_i = 1$ entails it
should.  We use the extractive model from
Section~\ref{sec:extract_sum} to produce probability distributions for
latent variables (see Equation~\eqref{eq:lbl_dist}) and obtain them by
sampling $z_i \sim p(z_i|z_{1:i-1}, \mathbf{h}^D_{i-1})$ (see
Figure~\ref{fig:model-overview}).
$\mathcal{C}=\{S_i | z_i = 1\}$, the set of sentences whose latent
variables equal to one, are our current extractive summaries. Without
loss of generality, we denote
$\mathcal{C}=(C_1,\dots,C_{|\mathcal{C}|})$. Then, we estimate how
likely it is to infer the human summary~$\mathcal{H}$
from~$\mathcal{C}$. We estimate the likelihood of summary sentence
$H_l$ given document sentence~$C_k$ with the compression model
introduced in Section~\ref{sec:compress} and calculate the
normalized\footnote{We also experimented with unnormalized
  probabilities (i.e., excluding the $\exp$ in Equation~\eqref{eq:s}),
  however we obtained inferior results.}  probability~$s_{kl}$:
\begin{equation}
\label{eq:s}
s_{kl} = \exp \left ( \frac{1}{|H_l|} \log p_{s2s}(H_l | C_k) \right )
\end{equation}
The score~$R_{p}$ measures the extent to which~$\mathcal{H}$ can be
inferred from~$\mathcal{C}$:
\begin{equation}
R_{p}(\mathcal{C}, \mathcal{H}) = \frac{1}{|\mathcal{C}|} \sum_{k=1}^{|\mathcal{C}|} \max_{l=1}^{|\mathcal{H}|} s_{kl}
\end{equation}
For simplicity, we assume one document sentence can only find one
summary sentence to explain it. Therefore, for all $H_l$, we only
retain the most evident $s_{kl}$.  $R_{p}(\mathcal{C}, \mathcal{H})$
can be viewed as the ``precision'' of document sentences with regard
to summary sentences. Analogously, we also define $R_{r}$, which
indicates the extent to which~$\mathcal{H}$ can be covered
by~$\mathcal{C}$:
\begin{equation}
R_{r}(\mathcal{C}, \mathcal{H}) = \frac{1}{|\mathcal{H}|} \sum_{l=1}^{|\mathcal{H}|} \max_{k=1}^{|\mathcal{C}|} s_{kl}
\end{equation}
$R_{r}(\mathcal{C}, \mathcal{H})$ can be viewed as the ``recall'' of
document sentences with regard to summary sentences. The final score
$R(\mathcal{C}, \mathcal{H})$ is the weighted sum of the two:
\begin{equation}
\label{eq:reward}
R(\mathcal{C}, \mathcal{H}) = \alpha \, R_{p}(\mathcal{C}, \mathcal{H}) + (1 - \alpha) \, R_{r}(\mathcal{C}, \mathcal{H})
\end{equation}
Our use of the terms ``precision'' and ``recall'' is reminiscent of
relevance and coverage in other summarization work
\cite{carbonell:1998:sigir,lin:2010:naacl,see:2017:acl}.

We train the model by minimizing the negative expected $R(\mathcal{C},
\mathcal{H})$:
\begin{equation}
\mathcal{L}(\theta) = - \mathbb{E}_{ (z_1,\dots,z_{|\mathcal{D}|}) \sim p(\cdot|\mathcal{D})} [ R(\mathcal{C}, \mathcal{H}) ]
\end{equation}
where $p(\cdot|\mathcal{D})$ is the distribution produced by the
neural extractive model (see
Equation~\eqref{eq:lbl_dist}). Unfortunately, computing the
expectation term is prohibitive, since the possible latent variable
combinations are exponential. In practice, we approximate this
expectation with a single sample from the distribution of
$p(\cdot|\mathcal{D})$.  We use the REINFORCE algorithm
\cite{williams:1992:ml} to approximate the gradient of
$\mathcal{L}(\theta)$:
\[
\begin{array}{ll}
\nabla\mathcal{L}(\theta)\approx \\
\sum_{i=1}^{|\mathcal{D}|}\nabla \log  p(z_i|z_{1:i-1}, \mathbf{h}^D_{i-1}) [R(\mathcal{C}, \mathcal{H}) - b_i ]
\end{array}
\]
Note that the model described above can be viewed as a reinforcement
learning model, where $R(\mathcal{C}, \mathcal{H})$ is the reward.  To
reduce the variance of gradients, we also introduce a baseline linear
regression\footnote{The linear regression model~$b_t$ is trained by
  minimizing the mean squared error between the prediction of $b_t$
  and $R(\mathcal{C}, \mathcal{H})$.} model $b_i$
\cite{ranzato:2016:iclr} to estimate the expected value of
$R(\mathcal{C}, \mathcal{H})$.  To avoid random label sequences during
sampling, we use a pre-trained extractive model to initialize our
latent model.

\section{Experiments}
\label{sec:exp}

\paragraph{Dataset and Evaluation}

We conducted experiments on the CNN/Dailymail dataset
\cite{hermann:2015:nips,see:2017:acl}. We followed the same
pre-processing steps as in \newcite{see:2017:acl}. The resulting
dataset contains 287,226 document-summary pairs for training, 13,368
for validation and 11,490 for test. To create sentence level labels,
we used a strategy similar to \newcite{nallapati:2017:aaai}. We label
the subset of sentences in a document that maximizes \textsc{Rouge}
(against the human summary) as {\tt True} and all other sentences as
{\tt False}. Using the method described in Section~\ref{sec:compress},
we created a compression dataset with~1,045,492 sentence pairs for
training, 53,434~for validation and 43,382~for testing.
We evaluated our models using full length F1 \textsc{Rouge}
\cite{lin:2004:acl:w} and the official {\tt ROUGE-1.5.5.pl} script. We
report \textsc{Rouge-1}, \textsc{Rouge-2}, and \textsc{Rouge-L}.

\paragraph{Implementation}
We trained our extractive model on an Nvidia K80 GPU card with a batch
size of 32. Model parameters were uniformly initialized
to~$[-\frac{1}{\sqrt{c}}, \frac{1}{\sqrt{c}}]$ ($c$ is the number of
columns in a weight matrix).  We used Adam \cite{kingma:2014:arxiv} to
optimize our models with a learning rate of~0.001, $\beta_1=0.9$,
and~$\beta_2=0.999$.
We trained our extractive model for 10~epochs and selected the model
with the highest \textsc{Rouge} on the validation set. We rescaled the
gradient when its norm exceeded~5 \cite{pascanu:2013:icml} and
regularized all LSTMs with a dropout rate of~0.3
\cite{srivastava:2014:jmlr,zaremba:2014:arxiv}.  We also applied word
dropout \cite{iyyer:2015:acl} at rate~0.2. We set the hidden unit size
$d=300$ for both word-level and sentence-level LSTMs and all LSTMs had
one layer. We used 300~dimensional pre-trained FastText vectors
\cite{joulin:2017:eacl} to initialize our word embeddings. The latent
model was initialized from the extractive model (thus both models have
the same size) and we set the weight in Equation~\eqref{eq:reward}
to~$\alpha=0.5$. The latent model was trained with SGD, with learning
rate~0.01 for 5 epochs. During inference, for both extractive and
latent models, we rank sentences with $p(y_i = \text{\tt True} |
y_{1:i-1}, \mathcal{D})$ and select the top three as summary (see also
Equation~\eqref{eq:lbl_dist}).

\paragraph{Comparison Systems}
We compared our model against {\sc Lead3}, which selects the first
three leading sentences in a document as the summary and a variety of
abstractive and extractive models. Abstractive models include a
sequence-to-sequence architecture \cite{nallapati:2016:arxiv};
\emph{abstract}), its pointer generator variant
(\citealt{see:2017:acl}; {\it pointer+coverage}), and two
reinforcement learning-based models (\citealt{paulus:2017:arxiv}; {\it
  abstract-RL} and {\it abstract-ML+RL}).  We also compared our
approach against an extractive model based on hierarchical recurrent
neural networks (\citealt{nallapati:2017:aaai}; {\it SummaRuNNer}),
the model described in
Section~\ref{sec:extract_sum} ({\sc extract}) which encodes sentences
using LSTMs, a variant which employs CNNs instead
(\citealt{cheng:2016:acl};  \textsc{extract-cnn}), as well as a
similar system based on reinforcement learning
(\citealt{Narayan:ea:2018}; \textsc{Refresh}).


%


\begin{table}
	\centering
	\small
	\begin{tabular}[t]{|@{~}l @{~}|@{~}c c c@{~}|}
          \hline
          Model & R-1 & R-2 & R-L \\
          \hline
          \hline
          {\sc Lead3} & 40.34 & 17.70 & 36.57 \\
          {\sc Lead3} \cite{nallapati:2017:aaai} & 39.20 & 15.70 &
          35.50 \\
          {\it abstract} & 35.46 & 13.30 & 32.65 \\
          {\it pointer+coverage} & 39.53 & 17.28 & 36.38 \\
          {\it abstract-RL}  & {\bf 41.16} & 15.75 & {\bf 39.08} \\
          {\it abstract-ML+RL} & 39.87 & 15.82 & 36.90 \\ \hline\hline
          {\it SummaRuNNer} & 39.60 & 16.20 & 35.30 \\
          {\sc extract-cnn} & 40.11 & 17.52 & 36.39 \\
          {\sc Refresh} \cite{Narayan:ea:2018} & 40.00 & 18.20 & 36.60 \\
          {\sc extract} & 40.62 & 18.45 & 37.14 \\
          {\sc latent} & 41.05 & {\bf 18.77} & 37.54 \\
          {\sc latent+compress} & 36.69 & 15.43 & 34.33 \\
          \hline
	\end{tabular}
	\caption{Results of different models on the CNN/Dailymail test
          set using full-length F1 \mbox{\textsc{Rouge-1}} (R-1),
          \textsc{Rouge-2} (R-2), and \textsc{Rouge-L} (R-L).}

	\label{tbl:cnndaily}
\end{table}

\paragraph{Results}
As shown in Table~\ref{tbl:cnndaily}, {\sc extract}, our extractive
model outperforms {\sc Lead3} by a wide margin. {\sc extract} also
outperforms previously published extractive models (i.e., {\it
  SummaRuNNer}, {\sc extract-cnn}, and {\sc refresh}). However, note
that {\it SummaRuNNer} generates anonymized summaries
\cite{nallapati:2017:aaai} while our models generate non-anonymized
ones, and therefore the results of {\sc extract} and {\it SummaRuNNer}
are not strictly comparable (also note that \textsc{Lead3} results are
different in Table \ref{tbl:cnndaily}). Nevertheless, {\sc extract}
exceeds {\sc lead3} by~$+0.75$ \mbox{\textsc{Rouge-2}} points
and~$+0.57$ in terms of \textsc{Rouge-L}, while {\it SummaRuNNer}
exceeds {\sc lead3} by~$+0.50$ \textsc{Rouge}-2 points and is worse
by~$-0.20$ points in terms of \textsc{Rouge-L}. We thus conclude that
{\sc extract} is better when evaluated with \textsc{Rouge-2} and
\textsc{Rouge-L}. {\sc extract} outperforms all abstractive models
except for {\it abstract-RL}. \textsc{Rouge}-2 is lower for {\it
  abstract-RL} which is more competitive when evaluated against
\textsc{Rouge-1} and \textsc{Rouge-l}.

Our latent variable model ({\sc latent}; Section \ref{sec:latent})
outperforms {\sc extract}, despite being a strong baseline, which
indicates that training with a loss directly based on gold summaries
is useful. Differences among {\sc Lead3}, {\sc extract}, and {\sc
  latent} are all significant with a~0.95 confidence interval
(estimated with the \textsc{Rouge} script).
Interestingly, when applying the compression model from
Section~\ref{sec:compress} to the output of our latent model (\mbox{ {\sc
  latent+compress} }), performance drops considerably. This may be
because the compression model is a sentence level model and it removes
phrases that are important for creating the document-level summaries.


\section{Conclusions}
We proposed a latent variable extractive summarization model which
leverages human summaries directly with the help of a sentence
compression model.  Experimental results show that the proposed model
can indeed improve over a strong extractive model while application of
the compression model to the output of our extractive system leads to
inferior output. In the future, we plan to explore ways to train
compression models tailored to our summarization task.

\section*{Acknowledgments}
We thank the EMNLP reviewers for their valuable feedback and Qingyu
Zhou for preprocessing the CNN/Dailymail dataset. We gratefully
acknowledge the financial support of the European Research Council
(award number 681760; Lapata).

\bibliography{emnlp2018}
\bibliographystyle{acl_natbib_nourl}

\end{document}